\documentclass{article}

\PassOptionsToPackage{numbers}{natbib}
\usepackage[preprint]{neurips_2019}

\usepackage[utf8]{inputenc} 
\usepackage[T1]{fontenc}    
\usepackage{hyperref}       
\usepackage{url}            
\usepackage{booktabs}       
\usepackage{amsfonts}       
\usepackage{nicefrac}       
\usepackage{microtype}      
\usepackage{graphicx}       
\usepackage{subcaption}

\usepackage{ifthen}
\usepackage{color}
\usepackage{glossaries}
\newboolean{showNotes} 
\usepackage{multirow}
\usepackage{authblk}

\setboolean{showNotes}{true}
\newcommand{\todo}[1]{\ifthenelse{\boolean{showNotes}}{\textcolor{red}{\textbf{\textcolor{red}{(TODO: #1)}}}}{}}

\title{An Empirical Study of Batch Normalization and Group Normalization in Conditional Computation}

\makeatletter
\renewcommand\AB@affilsepx{ \quad \protect\Affilfont}
\makeatother

\author[1,5]{Vincent Michalski}
\author[1,5]{Vikram Voleti}
\author[2,5]{Samira Ebrahimi Kahou}
\author[3]{Anthony Ortiz}
\author[1,5]{\authorcr Pascal Vincent}
\author[4,5,6]{Chris Pal}
\author[2,5]{Doina Precup}
\affil[1]{\footnotesize Université de Montréal}
\affil[2]{\footnotesize McGill University}
\affil[3]{\footnotesize University of Texas - El Paso\authorcr }
\affil[4]{\footnotesize Polytechnique Montréal}
\affil[5]{\footnotesize Quebec Artificial Intelligence Institute (Mila)}
\affil[6]{\footnotesize Element AI}

\begin{document}
\newacronym{bn}{BN}{Batch Normalization}
\newacronym{cbn}{CBN}{Conditional Batch Normalization}
\newacronym{cgn}{CGN}{Conditional Group Normalization}
\newacronym{cin}{CIN}{Conditional Instance Normalization}
\newacronym{clevr}{CLEVR}{Compositional Language and Elementary Visual Reasoning}
\newacronym{cln}{CLN}{Conditional Layer Normalization}
\newacronym{cogent}{CLEVR-CoGenT}{CLEVR Compositional Generalization Test}
\newacronym{fc100}{FC100}{Fewshot-CIFAR100}
\newacronym{fid}{FID}{Fr\'echet Inception Distance}
\newacronym{figureqa}{FigureQA}{Figure Question Answering}
\newacronym{film}{FiLM}{Feature-wise Linear Modulation}
\newacronym{gan}{GAN}{Generative Adversarial Network}
\newacronym{gn}{GN}{Group Normalization}
\newacronym{gru}{GRU}{gated recurrent unit}
\newacronym{in}{IN}{Instance Normalization}
\newacronym{is}{IS}{Inception Score}
\newacronym{ln}{LN}{Layer Normalization}
\newacronym{mse}{MSE}{mean-squared error}
\newacronym{mlp}{MLP}{multilayer perceptron}
\newacronym{nmn}{NMN}{Neural Module Network}
\newacronym{sagan}{SAGAN}{Self-Attention GAN}
\newacronym{sqoop}{SQOOP}{Spatial Queries On Object Pairs}
\newacronym{tadam}{TADAM}{Task dependent adaptive metric}
\newacronym{ten}{TEN}{task embedding network}
\newacronym{vqa}{VQA}{visual question answering}
\newacronym{cas}{CAS}{Classification Accuracy Score}

\maketitle

\begin{abstract}
  Batch normalization has been widely used to improve optimization in deep neural networks. While the uncertainty in batch statistics can act as a regularizer, using these dataset statistics specific to the training set impairs generalization in certain tasks. Recently, alternative methods for normalizing feature activations in neural networks have been proposed. Among them, group normalization has been shown to yield similar, in some domains even superior performance to batch normalization. All these methods utilize a learned affine transformation after the normalization operation to increase representational power. Methods used in conditional computation define the parameters of these transformations as learnable functions of conditioning information. In this work, we study whether and where the conditional formulation of group normalization can improve generalization compared to conditional batch normalization. We evaluate performances on the tasks of visual question answering, few-shot learning, and conditional image generation.
\end{abstract}

\section{Introduction}
\label{sec:intro}
In machine learning, the parameters of a model are typically optimized using a fixed training set. The model is then evaluated on
a separate partition of the data to estimate its generalization capability.
In practice, even under the i.i.d. assumption\footnote{All data samples are assumed to be drawn independently from an identical distribution (i.i.d.).}, the distribution of these two finite sets can \emph{appear} quite different to the learning algorithm, making it challenging to achieve strong and robust generalization. 
This difference is often the result of the fact that a training set of limited size cannot adequately cover the cross-product of all relevant factors of variation.
This issue can be addressed by making strong assumptions that simplify discovering a family of patterns from limited data. 
\citet{bahdanau2018systematic}, for example, show that their proposed synthetic relational reasoning task can be solved by a \gls{nmn}~\citep{andreas2016neural} with fixed tree structure, while models without this structural prior fail. 

Recent studies propose different benchmarks for evaluating task specific models for their generalization capacity~\citep{johnson2017clevr,kahou2017figureqa,bahdanau2018systematic}. 
While in this paper, we focus on \gls{vqa}, few-shot learning and generative models, any improvement in this direction can also benefit other domains such as reinforcement learning.  
Some of the best-performing models for each of these tasks are deep neural networks that employ \gls{cbn}~\citep{de2017modulating} for modulating normalized activations with contextual information. For \gls{bn}, one usually has to precompute activation statistics over the training set to be used during inference. 
Since \gls{bn}~\citep{ioffe2015batch} (and thus also \gls{cbn}) relies on dataset statistics, it seems that it may be vulnerable to significant domain shifts between training and test data. A recent study by \citet{galloway2019batch} indicates that \gls{bn} is also vulnerable to adversarial examples.

The recently proposed \gls{gn}~\citep{wu2018group} normalizes across groups of feature maps instead of across batch samples. 
Here, we explore whether a conditional formulation of \gls{gn} is a viable alternative for \gls{cbn}.
\Gls{gn} is conceptually simpler than \gls{bn}, as its function is the same during training and inference. 
Further, \gls{gn} can be used with small batch sizes, which may help in applications with particularly large feature maps, such as medical imaging or video processing, in which the available memory can be a constraint.

We compare \gls{cgn} and \gls{cbn} in a variety of tasks to see whether there are any significant performance differences. Section~\ref{sec:background} reviews some basic concepts that our work builds upon. Section~\ref{sec:experiments} describes setup and results of our experiments. Finally, we draw conclusions and present some directions for future work in Section~\ref{sec:conclusion}.

\section{Background}
\label{sec:background}
\subsection{Normalization Layers}
Several normalization methods have been proposed to stabilize and speed-up the training of deep neural networks~\citep{ioffe2015batch,wu2018group,lei2016layer,Ulyanov2016InstanceNT}. 
To stabilize the range of variation of network activations $x_i$, methods such as \gls{bn}~\citep{ioffe2015batch} first normalize the activations by subtracting mean $\mu_i$ and dividing by standard deviation $\sigma_i$:
\begin{equation}
    \label{eq:normalization}
    \hat x_i = \frac{1}{\sigma_i}\left(x_i - \mu_i\right)
\end{equation}
The distinction between different methods lies in how exactly these statistics are being computed. \citet{wu2018group} aptly summarize several methods using the following notation. Let $i=(i_N, i_C, i_H, i_W)$ be a four-dimensional vector, whose elements index the features along the batch, channel, height and width axes, respectively. The computation of the statistics can then be written as
\begin{equation}
    \label{eq:mu_sigma}
    \mu_i = \frac{1}{m}\sum\limits_{k\in\mathcal{S}_i} x_k, \;\;\; \sigma_i = \sqrt{\frac{1}{m}\sum\limits_{k\in\mathcal{S}_i} \left(x_k - \mu_i\right)^2 + \epsilon},
\end{equation}
where the set $\mathcal{S}_i$ of size $m$ is defined differently for each method and $\epsilon$ is a small constant for numerical stability.
\Gls{bn}, for instance, corresponds to: 
\begin{equation}
    \label{eq:bn_set}
    \text{BN} \implies \mathcal{S}_i = \{k \vert k_C = i_C\},
\end{equation}
i.e. $\mathcal{S}_i$ is the set of all pixels sharing the same channel axis, resulting in $\mu_i$ and $\sigma_i$ being computed along the $(N, H, W)$ axes.

As \citet{lei2016layer} point out, the performance of \gls{bn} is highly affected by the batch size hyperparameter. This insight led to the introduction of several alternative normalization schemes, that normalize per sample, i.e. not along batch axis $N$. \Gls{ln}~\citep{lei2016layer}, which normalizes activations within each layer, corresponds to the following set definition:
\begin{equation}
    \label{eq:ln_set}
    \text{LN} \implies \mathcal{S}_i = \{k \vert k_N = i_N\}.
\end{equation}
\citet{Ulyanov2016InstanceNT} introduce \gls{in} in the context of image stylization. \Gls{in} normalizes separately for each sample and each channel along the spatial dimensions: 
\begin{equation}
    \label{eq:in_set}
    \text{IN} \implies \mathcal{S}_i = \{k \vert k_N = i_N, k_C = i_C\}.
\end{equation}
Recently, \citet{wu2018group} introduced \gls{gn}, which draws inspiration from classical features such as HOG~\citep{dalal2005histograms}. It normalizes features per sample, separately within each of $G$ groups, along the channel axis:
\begin{equation}
    \label{eq:gn_set}
    \text{GN} \implies \mathcal{S}_i = \{k \vert k_N = i_N,
    \lfloor\frac{k_C}{C/G}\rfloor = \lfloor\frac{i_C}{C/G}\rfloor\}
\end{equation}
\Gls{gn} can be seen as a way to interpolate between the two extremes of \gls{ln} (corresponding to $G=1$, i.e. all channels are in a single group) and \gls{in} (corresponding to $G=C$, i.e. each channel is in its own group).

After normalization, all above mentioned methods insert a scaling and shifting operation using learnable per-channel parameters $\gamma$ and $\beta$:
\begin{equation}
    \label{eq:denormalization}
    y_i = \gamma\hat x_i + \beta
\end{equation}
This ``de-normalization'' is done to restore the representational power of the normalized network layer~\citep{ioffe2015batch}.

\Gls{cbn}~\citep{de2017modulating,perez2018film} is a conditional variant of \gls{bn}, in which the learnable parameters $\gamma$ and $\beta$ in Equation~\ref{eq:denormalization} are replaced by learnable functions
\begin{equation}
    \gamma(c_k) = W_\gamma c_k+b_\gamma,\,\,\, \beta(c_k) = W_\beta c_k + b_\beta
\end{equation}
of some per-sample conditioning input $c_k$ to the network with parameters $W_\gamma$, $W_\beta$, $b_\gamma$, $b_\beta$.
In a \gls{vqa} model, $c_k$ would for instance be an embedding of the question~\citep{perez2018film}.
\citet{Dumoulin2017ALR} introduce \gls{cin}, a conditional variant of \gls{in} similar to \gls{cbn}, replacing \gls{bn} with \gls{in}. 
In our experiments, we also explore a conditional variant of \gls{gn}.

\subsection{Visual Question Answering}
In \gls{vqa}~\citep{malinowski2014multi,antol2015vqa}, the task is to answer a question about an image. This task is usually approached by feeding both image and question to a parametric model, which is trained to predict the correct answer, for instance via classification among all possible answers in the dataset. 
One recent successful model for \gls{vqa} is the \gls{film} architecture~\citep{perez2018film}, which employs \gls{cbn} to modulate visual features based on an embedding of the question.

\subsection{Few-Shot Classification}
The task of few-shot classification consists in the challenge of classifying data given only a small set of support samples for each class.
In episodic $M$-way, $k$-shot classification tasks, meta-learning models~\citep{ravi2016optimization} learn to adapt a classifier given multiple $M$-class classification tasks, with $k$ support samples for each class. The meta-learner thus has to solve the problem of generalizing between these tasks given the limited number of training samples. In this work we experiment with the recently proposed \gls{tadam} architecture~\citep{oreshkin2018tadam}. It belongs to the family of meta-learners, that employ nearest neighbor classification within a learned embedding space. 
In the case of \gls{tadam}, the network providing this embedding is modulated by a task embedding using \gls{cbn}.

\subsection{Conditional Image Generation}
Some of the most successful models for generating images are \glspl{gan}~\citep{Goodfellow2014GenerativeAN}. 
This approach involves training a neural network (Generator) to generate an image, while the only supervisory signal is that from another neural network (Discriminator) which indicates whether the image looks real or not.
Several variants of \glspl{gan}~\citep{Mirza2014ConditionalGA, Odena2017ConditionalIS} have been proposed to condition the image generation process on a class label. More recently, the generators that work best stack multiple ResNet-style~\citep{he2016deep} architectural blocks, involving two CBN-ReLU-Conv operations and an upsampling operation. These blocks are followed by a BN-ReLU-Conv operation to transform the last features into the shape of an image.

Such models can be trained as Wasserstein \glspl{gan}
using gradient penalty (WGAN-GP) as proposed by \citet{Gulrajani2017ImprovedTO}, which gives mathematically sound arguments for an optimization framework. We adopt this framework for our experiments. 
More recently, two of the most noteworthy \gls{gan} architectures, \gls{sagan}~\citep{Han18sagan} and BigGAN~\citep{Brock2019LargeSG}, use architectures similar to WGAN-GP, with some important changes. \Gls{sagan} inserts a self-attention mechanism~\citep{parikh2016decomposable,vaswani2017attention,cheng2016long} to attend over important parts of features during the generation process. In addition, it uses spectral normalization~\citep{miyato2018spectral} to stabilize training. The architecture of BigGAN is the same as for \gls{sagan}, with the exception of an increase in batch size and channel widths, as well as some architectural changes to improve memory and computational efficiency. Both these models have been successfully used in generating high quality natural images. In our experiments, we compare performance metrics of WGAN-GP networks using two types of normalization.

\section{Experiments}
\label{sec:experiments}
\subsection{Visual Question Answering}
\label{sec:exp-vqa}
We study whether substituting \gls{cgn} for \gls{cbn} in the \gls{vqa} architecture \gls{film}~\citep{perez2018film} yields comparable performance. We run experiments on several recently proposed benchmarks for compositional generalization.

\subsubsection{Datasets}
\begin{figure}
\begin{subfigure}[t]{0.3\textwidth}
    \centering
    \includegraphics[width=1\textwidth]{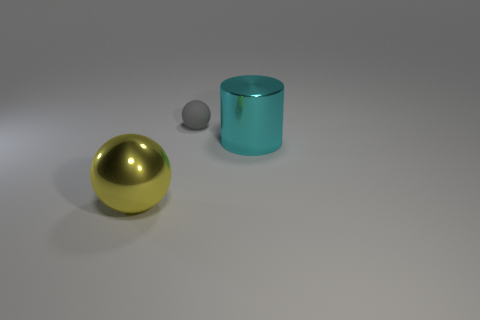}
    \caption{\textbf{\gls{cogent}}: Are there any gray things made of the same material as the big cyan cylinder? - No}
    \label{fig:cogent}
\end{subfigure}
\hfill
\begin{subfigure}[t]{0.3\textwidth}
    \centering
    \includegraphics[width=1\textwidth]{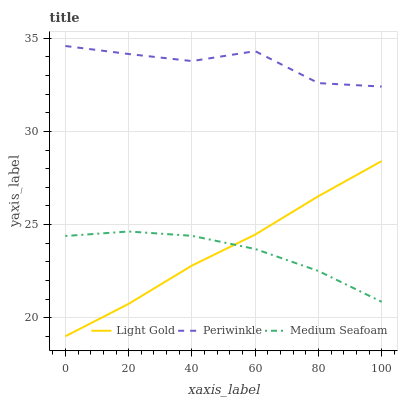}
    \caption{\textbf{\gls{figureqa}}: Does Medium Seafoam intersect Light Gold? - Yes}
    \label{fig:figureqa}
\end{subfigure}
\hfill
\begin{subfigure}[t]{0.3\textwidth}
    \centering
    \includegraphics[width=1\textwidth]{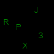}
    \caption{\textbf{\gls{sqoop}}: X right\_of J? - no}
    \label{fig:sqoop}
\end{subfigure}
\caption{Examples of the \gls{vqa} datasets used in our experiments.}
\label{fig:data_samples_vqa}
\end{figure}

\textbf{\gls{cogent}}~\citep{johnson2017clevr} is a variant of the popular \gls{clevr} dataset~\citep{johnson2017clevr}, that tests for compositional generalization. The images consist of rendered three-dimensional scenes containing several shapes (small and large cubes, spheres and cylinders) of differing material properties (\emph{metal} or \emph{rubber}) and colors. Questions involve \textit{queries} for object attributes, \textit{comparisons}, \textit{counting} of sets and combinations thereof. In contrast to the regular \gls{clevr} dataset, the training set of \gls{cogent} explicitly combines some shapes only with different subsets of four out of eight colors, and provides two validation sets: one with the same combinations (\textit{valA}) and one in which the shape-color assignments are swapped (\textit{valB}). To perform well on \textit{valB}, the model has to generalize to unseen combinations of shapes and colors, i.e. it needs to somewhat capture the compositionality of the task.
Figure~\ref{fig:cogent} shows an example from this dataset.

\textbf{\gls{figureqa}}~\citep{kahou2017figureqa} is a \gls{vqa} dataset consisting of mathematical plots with templated yes/no question-answer pairs that address relations between plot elements. The dataset contains plots of five types (vertical/horizontal bar plots, line plots, pie charts and dot-line plots). Each plot has between 2 and 10 elements, each of which has one of 100 colors.
Plot elements (e.g. a slice in a pie chart) are identified by their color names in the questions. Questions query for \textit{one-vs-one} or \textit{one-vs-all} attribute relations, e.g. "Is Lime Green less than WebGray?" or "Does Cadet Blue have the minimum area under the curve?". 
Similar to \gls{cogent}, \gls{figureqa} requires compositional generalization. 
The overall 100 colors are split into two sets $A$ and $B$, each containing 50 unique colors. 
During training, colors of certain plot types are sampled from set $A$, while the remaining plot types use colors from set $B$ (\emph{scheme 1}). There are two validation sets, one using the same color scheme, and one for which the plot-type to color assignments are swapped (\emph{scheme 2}). See Figure~\ref{fig:figureqa} for a sample from the dataset.

\textbf{\gls{sqoop}}~\citep{bahdanau2018systematic} is a recently introduced dataset that tests for systematic generalization. 
It consists of images containing five randomly chosen and arranged objects (digits and characters). 
Questions concern the four spatial relations \textit{LEFT OF}, \textit{RIGHT OF}, \textit{ABOVE} and \textit{BELOW} and the queries are all of the format "X R Y?", where X and Y are left-hand and right-hand objects and R is a relationship between them, e.g. "nine LEFT OF a?". 
To test for systematic generalization, only a limited number of combinations of each left-hand object with different right-hand objects Y are shown during training. In the hardest version of the task (1 rhs/lhs), only a single right-hand side object is combined with each left-hand side object. For instance, the training set of this version may contain images with the query "A RIGHT OF B", but no images with queries about relations of left-hand object A with any other object than B.
The test set contains images and questions about all combinations, i.e. it evaluates generalization to relations between novel object combinations.
Figure~\ref{fig:sqoop} shows an example from the training set.

\subsubsection{Model}
We experiment with several small variations of the \gls{film} architecture~\citep{perez2018film}.
The original architecture in \citet{perez2018film} consists of an unconditional \emph{stem} network, a core of four ResNet~\citep{he2016deep} blocks with \gls{cbn}~\citep{de2017modulating} and a classifier.
The stem network is either a sequence of residual blocks trained from scratch or a fixed pre-trained feature extractor followed by a learnable layer of $3\times3$ convolutions. The scaling and shifting parameters of the core layers are affine transforms of a question embedding provided by a \gls{gru}~\citep{cho2014properties}.
The output of the last residual block is fed to the classifier, which consists of a layer of 512 $1\times1$ convolutions, global max-pooling, followed by a fully-connected ReLU~\citep{nair2010rectified} layer using (unconditional) \gls{bn} and a softmax layer, which outputs the probability of each possible answer. 
We train the following three variants that include \gls{cgn}\footnote{We always set the number of groups to 4, as the authors of \citet{wu2018group} showed that this hyperparameter does not have a large influence on the performance. This number was selected using uniform sampling from the set $\{2,4,8,16\}$.}:
\begin{enumerate}
    \item all conditional and regular \gls{bn} layers are replaced with corresponding conditional or regular \gls{gn} layers.
    \item all \gls{cbn} layers are replaced with \gls{cgn}, regular \gls{bn} layers are left unchanged.
    \item all \gls{cbn} layers are replaced with \gls{cgn}, regular \gls{bn} layers are left unchanged, except the fully-connected hidden layer in the classifier, for which we remove normalization.
\end{enumerate}

Besides the described changes in the normalization layers, the architecture and hyperparameters are the same as used in \citet{perez2018film} for all experiments, except for \gls{sqoop} where they are the same as in \citet{bahdanau2018systematic}. The only difference is that we set the constant $\epsilon$ of the Adam optimizer~\citep{kingma2014adam} to $1\mathrm{e}{-5}$ to improve training stability\footnote{The authors of \citet{perez2018film} confirmed occasional gradient explosions with the original setting of $1\mathrm{e}-8$.}.
For \gls{sqoop}, the input to the residual network are the raw image pixels. For all other networks, the input is features extracted from layer \emph{conv4} of a ResNet-101~\citep{he2016deep}, pre-trained on ImageNet~\citep{russakovsky2015imagenet}, following \citet{perez2018film}.

\subsubsection{Results}
Tables~\ref{tab:vqa-results-cogent}, \ref{tab:vqa-results-figureqa} and \ref{tab:vqa-results-sqoop} show the results of training \gls{film} with \gls{cbn} and \gls{cgn} on the three considered datasets.
In the experiments on \gls{cogent}, all three \gls{cgn} variants of \gls{film} achieve a slightly higher average accuracy. 
On \gls{figureqa}, \gls{cbn} outperforms \gls{cgn} slightly.
In the hardest \gls{sqoop} variant with only one right-hand side object per left-hand side object (\emph{1 rhs/lhs}), all three variants of \gls{cgn} achieve a higher performance than \gls{cbn}. 
For \gls{sqoop} variants whose training sets contain more combinations, \gls{cgn} did not converge in some cases. 
Learning curves of models successfully trained on \gls{sqoop} seem to follow the same pattern: For a relatively large number of gradient updates there is no significant improvement. Then, at some point, almost instantly the model achieves $100\%$ training accuracy. It is possible that a hyperparameter search or additional regularization is required to guarantee convergence.

\begin{table}[ht!]
  \caption{Classification accuracy on \gls{cogent} \emph{valB}. Mean and standard deviation of three runs with early stopping on \emph{valA} are reported for the models we trained.}
  \label{tab:vqa-results-cogent}
  \centering
  \begin{tabular}{ll}
    \toprule
    Model & Accuracy (\%) \\
    \hline\vspace{-0.8em}\\
    \gls{cbn} (\gls{film}~\citep{perez2018film}) &  
    $75.600$ \\
    \gls{cbn} (\gls{film}, our results) &
    $75.539 \pm 0.671$ \\
    \gls{cgn} (all \gls{gn}) & 
    $75.758 \pm 0.356$ \\
    \gls{cgn} (\gls{bn} in stem, classifier no norm) & 
    $75.703 \pm 0.571$ \\
    \gls{cgn} (\gls{bn} in stem and classifier) & 
    $\mathbf{75.807 \pm 0.511}$ \\
    \bottomrule
  \end{tabular}
\end{table}

\begin{table}[ht!]
  \caption{Classification accuracy on \gls{figureqa} \emph{validation2}, mean and standard deviation of three runs after early stopping on \emph{validation1}.}
  \label{tab:vqa-results-figureqa}
  \centering
  \begin{tabular}{ll}
    \toprule
    Model & Accuracy (\%) \\
    \hline\vspace{-0.8em}\\
    \gls{cbn} (\gls{film}, our results) &  
    $\mathbf{91.618 \pm 0.132}$ \\
    \gls{cgn} (all \gls{gn}) & 
    $91.343 \pm 0.436$ \\
    \gls{cgn} (\gls{bn} in stem, classifier no norm) & 
    $91.080 \pm 0.166$ \\
    \gls{cgn} (\gls{bn} in stem and classifier) & 
    $91.317 \pm 0.514$ \\
    \bottomrule
  \end{tabular}
\end{table}

\begin{table}[ht!]
  \caption{Test accuracies on several versions of \gls{sqoop}. Mean and standard deviation of three runs after early stopping on the validation set are reported for the models we trained.}
  \label{tab:vqa-results-sqoop}
  \centering
  \begin{tabular}{lll}
    \toprule
    Dataset & Model & Accuracy (\%) \\
    \hline\vspace{-0.8em}\\
    \multirow{2}{*}{1 rhs/lhs} & 
    \gls{cbn} (\gls{film}~\citep{bahdanau2018systematic}) & $65.270 \pm 4.610$ \\
    & \gls{cbn} (\gls{film}, our results) & $72.369 \pm 0.529$ \\
    & \gls{cgn} (all \gls{gn}) & 
    $74.020 \pm 2.814$ \\
    & \gls{cgn} (\gls{bn} in stem, classifier no norm) & 
    $73.824 \pm 0.334$ \\
    & \gls{cgn} (\gls{bn} in stem and classifier) & 
    $\mathbf{74.929 \pm 3.888}$ \\
    \midrule
    \multirow{2}{*}{2 rhs/lhs} & 
    \gls{cbn} (\gls{film}~\citep{bahdanau2018systematic}) & $80.200 \pm 4.320$ \\
    & \gls{cbn} (\gls{film}, our results) & $84.966 \pm 4.165$ \\
    & \gls{cgn} (all \gls{gn}) & 
    $\mathbf{86.689 \pm 6.308}$ \\
    & \gls{cgn} (\gls{bn} in stem, classifier no norm) &
    $83.109 \pm 0.381$ \\
    & \gls{cgn} (\gls{bn} in stem and classifier) & 
    $85.859 \pm 5.318$ \\
    \midrule
    \multirow{2}{*}{4 rhs/lhs} & 
    \gls{cbn} (\gls{film}~\citep{bahdanau2018systematic}) & $90.420 \pm 1.000$ \\
    & \gls{cbn} (\gls{film}, our results) & $97.043 \pm 1.958$ \\
    & \gls{cgn} (all \gls{gn}) & 
    $91.404 \pm 0.318$ \\
    & \gls{cgn} (\gls{bn} in stem, classifier no norm) &
    $91.601 \pm 1.937$ \\
    & \gls{cgn} (\gls{bn} in stem and classifier) & 
    $\mathbf{99.474 \pm 0.254}$ \\
    \midrule
    \multirow{2}{*}{35 rhs/lhs} & 
    \gls{cbn} (\gls{film}~\citep{bahdanau2018systematic}) & $99.803 \pm 0.219$ \\
    & \gls{cbn} (\gls{film}, our results) & $\mathbf{99.841 \pm 0.043}$ \\ 
    & \gls{cgn} (all \gls{gn}) & 
    $99.755 \pm 0.025$ \\
    & \gls{cgn} (\gls{bn} in stem, classifier no norm) &
    $99.815 \pm  0.122$ \\
    & \gls{cgn} (\gls{bn} in stem and classifier) & 
    $99.782 \pm 0.155$ \\
    \bottomrule
  \end{tabular}
  \vspace{-1em}
\end{table}

\subsection{Few-Shot Learning}
\label{sec:exp-fewshot}
\Gls{cbn} has also been used in recent methods for few-shot learning~\citep{oreshkin2018tadam,jiang2018learning}. We replicate the experiments of \citet{oreshkin2018tadam} on Mini-ImageNet and \gls{fc100} using their code for \gls{tadam}\footnote{\url{https://github.com/ElementAI/TADAM}} and compare the results with a version that uses \gls{cgn} instead of \gls{cbn}.

\subsubsection{Datasets}
\textbf{Mini-ImageNet}
was proposed by \citet{vinyals2016matching} as a benchmark for few-shot classification. It contains 100 classes, for each of which there are 600 images of resolution $84\times84$. To generate five-way five-shot classification tasks five classes and five support samples for each class are sampled uniformly. The remaining images are used to compute the accuracy. 
Using the proposed split by \citet{ravi2016optimization}, we uniformly sample training tasks from a subset of 64 classes. The remaining 36 classes are divided into 16 for meta-validation and 20 for meta-testing. 

\textbf{Fewshot-CIFAR100}~\citep{oreshkin2018tadam} is a few-shot classification version of the popular CIFAR100 data set~\citep{cifar10}. Similarly to Mini-ImageNet, it contains 100 classes and 600 samples per class. The resolution of the images is $32\times32$. The classes are split by superclasses to reduce information overlap between data set partitions, which makes the task more challenging than Mini-ImageNet. The training partition contains 60 classes belonging to 12 superclasses. The validation and test partitions contain 20 classes belonging to 5 superclasses each. The tasks are sampled uniformly as in Mini-ImageNet.

\subsubsection{Model}
\begin{figure}
    \centering
    \includegraphics[width=\textwidth]{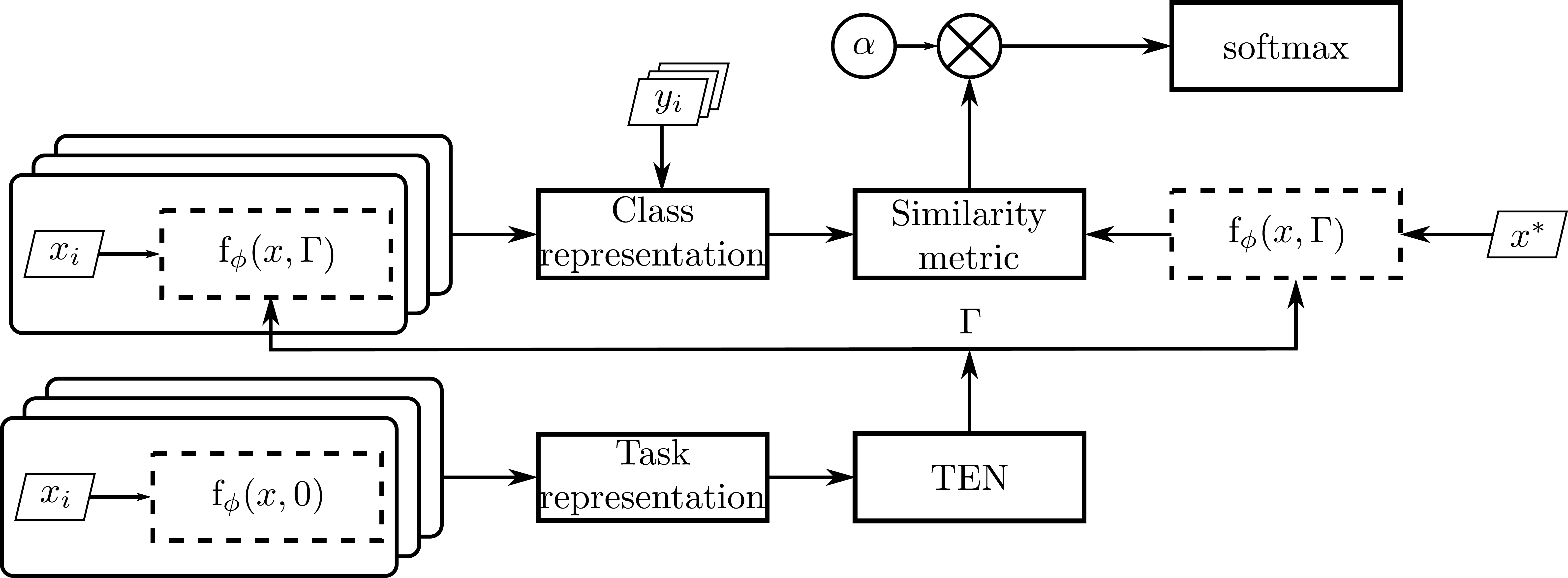}
    \caption{Architecture of \gls{tadam}~\citep{oreshkin2018tadam}. Boxes with dashed border share parameters. Figure adapted from \citep{oreshkin2018tadam}.}
    \label{fig:tadam}
\end{figure}
\Gls{tadam}~\citep{oreshkin2018tadam} is a metric-based few-shot classifier, i.e. it learns a measure of similarity between query samples and class representations. The metric is based on a learned image embedding $\mathrm{f_\phi}(x, c)$ provided by a residual network. 
Figure~\ref{fig:tadam} shows a diagram of the overall architecture.
Each class template is computed as the average embedding of all support samples for the respective class. The Euclidean distances between the embedding of a query sample and each of the class templates, weighted by a learned scaling factor $\alpha$, is then used to classify the query sample  $x^\ast$.
The embedding network $\mathrm{f_\phi}$ (see the dashed boxes in Figure~\ref{fig:tadam}) is modulated using \gls{cbn} with a conditioning input $c$.
In the computation of the similarity metric, $c$ is fed by a task embedding $\Gamma$ provided by a \acrfull{ten}, which reads the average embeddings of support samples from all classes of the task. Note that $\mathrm{f_\phi}$ is evaluated without conditioning (i.e. by setting $c$ to a zero vector\footnote{The conditioning input is implemented as a deviation from the identity transform (unity scaling and zero shift), so setting it to zero does not change the normalized activations.}) in the computation of the task embedding $\Gamma$ (see bottom of Figure~\ref{fig:tadam}).
For the \gls{gn} version we replaced all conditional and regular \gls{bn} layers with their corresponding conditional or regular \gls{gn} version (with the number of groups set to 4). For a complete description of the experimental setup, including all other hyperparameters, we refer the reader to \citet{oreshkin2018tadam}.

\subsubsection{Results} 
\begin{table}[ht!]
  \caption{Five-way five-shot classification accuracy on Fewshot-CIFAR100~\cite{oreshkin2018tadam} and Mini-Imagenet~\cite{vinyals2016matching}, mean and standard deviation of ten runs.}
  \label{tab:fewshot-results}
  \centering
  \begin{tabular}{lll}
    \toprule
    Dataset & Model & Accuracy (\%) \\
    \hline\vspace{-0.8em}\\
    \multirow{2}{*}{FC100} & TADAM (\gls{cbn})~\cite{oreshkin2018tadam} & $\mathbf{52.996 \pm 0.610}$ \\
    & TADAM (\gls{cgn}) & $52.807 \pm 0.509$ \\
    \midrule
    \multirow{2}{*}{Mini-Imagenet} & TADAM (\gls{cbn})~\cite{oreshkin2018tadam} & $\mathbf{76.414 \pm 0.499}$ \\
    & TADAM (\gls{cgn}) & $74.032 \pm 0.373$ \\
    \bottomrule
  \end{tabular}
\end{table}
We see that using \gls{cgn} instead of \gls{cbn} yields only slightly reduced performance on \gls{fc100}, while there is a considerable $2.4\%$ gap for Mini-ImageNet. Note, that we simply reuse the hyperparameters from \citet{oreshkin2018tadam}, which were tuned for \gls{cbn}.

\subsection{Conditional Image Generation}
\label{sec:exp-gan}
Here we compare \gls{cbn} and \gls{cgn} on the task of generating images conditioned on their class label using the WGAN-GP~\citep{Gulrajani2017ImprovedTO} architecture.

\subsubsection{Dataset}
CIFAR-10~\citep{cifar10} is a data set containing 60000 $32\times32$ images, 6000 for each of 10 classes. The dataset is split into 50000 training and 10000 test samples. 

\subsubsection{Model}
We replicated the WGAN-GP~\citep{Gulrajani2017ImprovedTO} architecture from the original paper, which uses \gls{cbn}. As in other tasks, we also train the \gls{cgn} variants, where we substitute conditional and unconditional \gls{bn} layers with the corresponding conditional or unconditional \gls{gn} layers, with number of groups set to 4. We use the optimization setup from \citet{Gulrajani2017ImprovedTO}: a learning rate of $2\mathrm{e}^{-4}$ for both generator and discriminator, five discriminator updates per generator update, and we also use the Adam optimizer~\citep{kingma2014adam}. We train using a single GPU (NVIDIA P100) and a batch size of 64.

\subsubsection{Results}
\begin{figure}
\begin{tabular}{cc}
  \includegraphics[width=.45\textwidth]{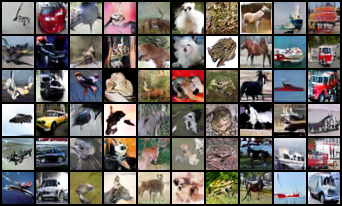} &   \includegraphics[width=.45\textwidth]{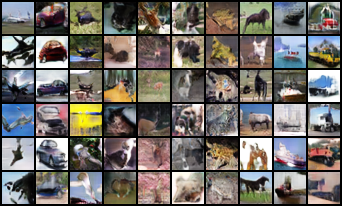} \\
(a) CBN & (b) CGN \\
\end{tabular}
\caption{Samples from models trained with different normalization techniques. The images in each column belong to the same class, ordered as `airplane', `automobile', `bird', `cat', `deer', `dog', `frog', `horse', `ship', `truck'. Samples are not cherry-picked.}
\label{fig:samples_WGANGP_CIFAR10}
\vspace{-1em}
\end{figure}

Figure~\ref{fig:samples_WGANGP_CIFAR10} shows samples from WGAN-GP trained using each of the two normalization methods.
For both normalization methods, in addition to a qualitative check of the generated samples, we calculate two scores that are widely used in the community to evaluate image generation \gls{is}~\citep{salimans2016improved} and \gls{fid}~\citep{Heusel2017GANsTB}. We use publicly available code to calculate \gls{is}\footnote{https://github.com/sbarratt/inception-score-pytorch} and \gls{fid}\footnote{https://github.com/mseitzer/pytorch-fid}. The computed values for real data may differ slightly from the original ones since these use PyTorch~\citep{paszke2017automatic} implementations, while the original papers use TensorFlow~\citep{tensorflow2015-whitepaper}. However, we compare the same implementation of these metrics for true and generated data.

\gls{is} is meant to measure the natural-ness of an image by checking the embedding of the generated images on a pre-trained Inception network~\citep{szegedy2016rethinking}. Although the suitability of the \gls{is} for this purpose has been rightfully put into question~\citep{Barratt2018ANO}, it continues to be used frequently.
\gls{fid} measures how similar two sets of images are, by computing the Fréchet distance between two multivariate Gaussians fitted to the embeddings of the images from the two sets. The embeddings are obtained from a pre-trained InceptionV3 network~\citep{szegedy2016rethinking}. In this case, we measure the distance between the real CIFAR-10 images, and the generated ones. This is a better metric than \gls{is}, since there is no constraint on the images being natural, and it is able to quantify not only their similarity to the real images, but also diversity in the generated images.

\begin{figure}
\begin{tabular}{cc}
  \includegraphics[width=.45\textwidth]{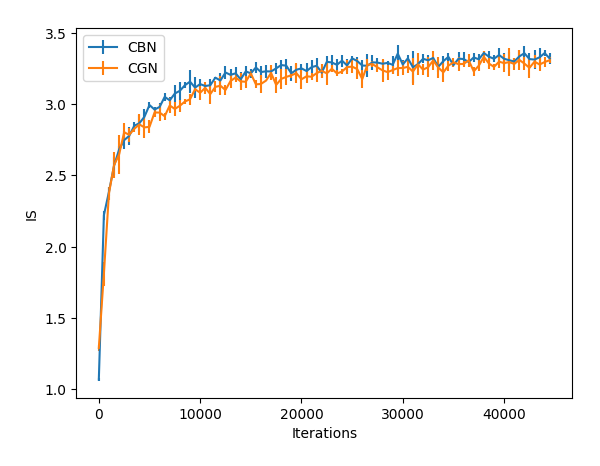} &   \includegraphics[width=.45\textwidth]{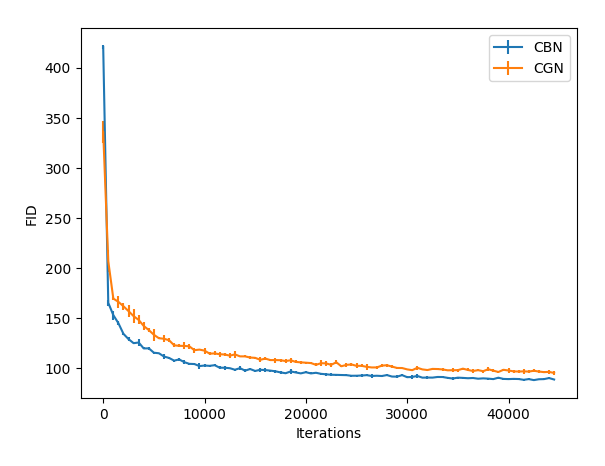} \\
(a) IS & (b) FID \\
\end{tabular}
\caption{(a) Inception score (IS, higher is better) and (b) FID (lower is better) of samples generated by WGAN-GP model while training on CIFAR-10.}
\label{fig:IS_FID_WGANGP_CIFAR10}
\vspace{-1em}
\end{figure}

We first calculate the IS of the true images of CIFAR-10, for each class separately. Then, during training of a model, we sample images from the generator at regular intervals, and calculate the \gls{is} and \gls{fid} of those images for each class separately. This allows us to see the effect of the different normalization techniques on the conditional generation process. We average our results from four runs with different seeds, shown in Figure~\ref{fig:IS_FID_WGANGP_CIFAR10}.

\begin{figure}[ht]
\begin{center}
\centerline{\includegraphics[width=.5\textwidth]{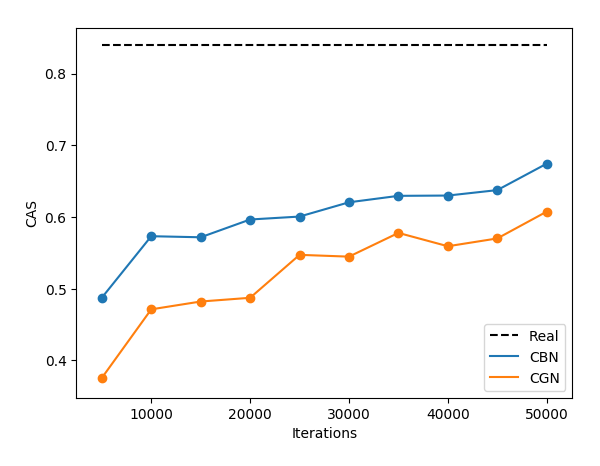}}
\vspace{-1em}
\caption{Classification Accuracy Score (CAS) using a ResNet classifier trained on samples generated while training on CIFAR-10 with WGAN-GP using (blue) CBN and (green) CGN, while (black) is the value when trained with true data. All classifiers have been trained with the same hyperparameters.}
\label{fig:CAS_WGANGP_CIFAR10}
\end{center}
\vspace{-2em}
\end{figure}

We also calculate the recently proposed \gls{cas}~\citep{ravuri2019cas} for one instance of training using WGAN-GP with \gls{cbn} and \gls{cgn} each, shown in Figure~\ref{fig:CAS_WGANGP_CIFAR10}. In the computation of this metric, a ResNet~\citep{he2016deep} classifier is trained on data sampled from the generative model being evaluated. Then the accuracy of this classifier on the true validation data is calculated. \citet{ravuri2019cas} mention that this could indicate the closeness of the generated data distribution to the true data distribution. All three metrics indicate that \gls{cbn} is better than \gls{cgn} in conditional generative models of images such as WGAN-GP.

The WGAN-GP model architecture consists of a series of residual blocks followed by bn-relu-conv layers. Each residual block contains two bn-relu-conv modules. Since the architectures of more recent models such as \gls{sagan}~\citep{Han18sagan} and BigGAN~\citep{Brock2019LargeSG} are similar to that of the one we used, it is likely that the conclusions we draw from the WGAN-GP experiments transfer to them.

\section{Conclusion}
\label{sec:conclusion}
Because the performance of \gls{cbn} heavily depends on the batch size and on how well training and test statistics match, we investigate the use of \gls{cgn} as a potential alternative for \gls{cbn}.
We consider a set of experiments for \gls{vqa}, few-shot learning and image generation tasks in which some of the best models rely on \gls{cbn} for conditional computation. 
We experimentally show that the effect of this substitution is task-dependent, with performance increases in some \gls{vqa} tasks that focus on systematic generalization, but a clear decrease in performance in conditional image generation. 
\Gls{cgn}'s simpler implementation, its consistent behaviour during training and inference time, as well as its independence from batch sizes, are all good reasons to explore its adoption instead of \gls{cbn} in tasks that require systematic generalization.
That being said, further analysis is required to be able to confidently suggest one method over the other.
For instance, a hyperparameter search for each of the normalization methods would be required to provide a better performance comparison. 
Also, we would like to characterize the sensitivity of \gls{cbn}'s performance to the batch size and focus on domains, such as medical imaging or video processing, for which efficient large-batch training becomes nontrivial.
Lastly, since some of the success of \gls{bn} (and consequently also \gls{cbn}) can be attributed to the regularization effect introduced by noisy batch statistics, it seems worthwile to explore combinations of \gls{cgn} with additional regularization as suggested for \gls{gn} by \citet{wu2018group}.
The latter is also motivated by recent successful attempts at replacing (unconditional) \gls{bn} with careful network initialization~\citep{zhang2018fixup}, which relies on additional regularization~\citep{zhang2018mixup} to match generalization performance.

\subsubsection*{Acknowledgments}
We thank Boris Oreshkin, Eugene Belilovsky, Matthew Scicluna, Mahdi Ebrahimi Kahou, Kris Sankaran and Alex Lamb for helpful discussions.
This research was enabled in part by support provided by Compute Canada.

\bibliographystyle{unsrtnat}
\bibliography{literature}

\begin{thebibliography}{44}
\providecommand{\natexlab}[1]{#1}
\providecommand{\url}[1]{\texttt{#1}}
\expandafter\ifx\csname urlstyle\endcsname\relax
  \providecommand{\doi}[1]{doi: #1}\else
  \providecommand{\doi}{doi: \begingroup \urlstyle{rm}\Url}\fi

\bibitem[Bahdanau et~al.(2018)Bahdanau, Murty, Noukhovitch, Nguyen, de~Vries,
  and Courville]{bahdanau2018systematic}
Dzmitry Bahdanau, Shikhar Murty, Michael Noukhovitch, Thien~Huu Nguyen, Harm
  de~Vries, and Aaron Courville.
\newblock Systematic generalization: What is required and can it be learned?
\newblock \emph{arXiv preprint arXiv:1811.12889}, 2018.

\bibitem[Andreas et~al.(2016)Andreas, Rohrbach, Darrell, and
  Klein]{andreas2016neural}
Jacob Andreas, Marcus Rohrbach, Trevor Darrell, and Dan Klein.
\newblock Neural module networks.
\newblock In \emph{Proceedings of the IEEE Conference on Computer Vision and
  Pattern Recognition (CVPR)}, pages 39--48, 2016.

\bibitem[Johnson et~al.(2017)Johnson, Hariharan, van~der Maaten, Fei-Fei,
  Zitnick, and Girshick]{johnson2017clevr}
Justin Johnson, Bharath Hariharan, Laurens van~der Maaten, Li~Fei-Fei,
  C~Lawrence Zitnick, and Ross Girshick.
\newblock Clevr: A diagnostic dataset for compositional language and elementary
  visual reasoning.
\newblock In \emph{Proceedings of the IEEE Conference on Computer Vision and
  Pattern Recognition (CVPR)}, 2017.

\bibitem[Kahou et~al.(2017)Kahou, Michalski, Atkinson, K{\'a}d{\'a}r,
  Trischler, and Bengio]{kahou2017figureqa}
Samira~Ebrahimi Kahou, Vincent Michalski, Adam Atkinson, {\'A}kos
  K{\'a}d{\'a}r, Adam Trischler, and Yoshua Bengio.
\newblock Figureqa: An annotated figure dataset for visual reasoning.
\newblock \emph{Workshop in the International Conference on Learning
  Representations}, 2017.

\bibitem[De~Vries et~al.(2017)De~Vries, Strub, Mary, Larochelle, Pietquin, and
  Courville]{de2017modulating}
Harm De~Vries, Florian Strub, J{\'e}r{\'e}mie Mary, Hugo Larochelle, Olivier
  Pietquin, and Aaron~C Courville.
\newblock Modulating early visual processing by language.
\newblock In \emph{Advances in Neural Information Processing Systems}, pages
  6594--6604, 2017.

\bibitem[Ioffe and Szegedy(2015)]{ioffe2015batch}
Sergey Ioffe and Christian Szegedy.
\newblock Batch normalization: Accelerating deep network training by reducing
  internal covariate shift.
\newblock In \emph{International Conference on Machine Learning (ICML)}, 2015.

\bibitem[Galloway et~al.(2019)Galloway, Golubeva, Tanay, Moussa, and
  Taylor]{galloway2019batch}
Angus Galloway, Anna Golubeva, Thomas Tanay, Medhat Moussa, and Graham~W
  Taylor.
\newblock Batch normalization is a cause of adversarial vulnerability.
\newblock \emph{arXiv preprint arXiv:1905.02161}, 2019.

\bibitem[Wu and He(2018)]{wu2018group}
Yuxin Wu and Kaiming He.
\newblock Group normalization.
\newblock In \emph{Proceedings of the European Conference on Computer Vision
  (ECCV)}, pages 3--19, 2018.

\bibitem[Lei~Ba et~al.(2016)Lei~Ba, Kiros, and Hinton]{lei2016layer}
Jimmy Lei~Ba, Jamie~Ryan Kiros, and Geoffrey~E Hinton.
\newblock Layer normalization.
\newblock \emph{arXiv preprint arXiv:1607.06450}, 2016.

\bibitem[Ulyanov et~al.(2016)Ulyanov, Vedaldi, and
  Lempitsky]{Ulyanov2016InstanceNT}
Dmitry Ulyanov, Andrea Vedaldi, and Victor~S. Lempitsky.
\newblock Instance normalization: The missing ingredient for fast stylization.
\newblock \emph{CoRR}, abs/1607.08022, 2016.

\bibitem[Dalal and Triggs(2005)]{dalal2005histograms}
Navneet Dalal and Bill Triggs.
\newblock Histograms of oriented gradients for human detection.
\newblock In \emph{international Conference on computer vision \& Pattern
  Recognition (CVPR'05)}, volume~1, pages 886--893. IEEE Computer Society,
  2005.

\bibitem[Perez et~al.(2018)Perez, Strub, De~Vries, Dumoulin, and
  Courville]{perez2018film}
Ethan Perez, Florian Strub, Harm De~Vries, Vincent Dumoulin, and Aaron
  Courville.
\newblock Film: Visual reasoning with a general conditioning layer.
\newblock In \emph{Thirty-Second AAAI Conference on Artificial Intelligence},
  2018.

\bibitem[Dumoulin et~al.(2017)Dumoulin, Shlens, and Kudlur]{Dumoulin2017ALR}
Vincent Dumoulin, Jonathon Shlens, and Manjunath Kudlur.
\newblock A learned representation for artistic style.
\newblock \emph{International Conference on Learning Representations (ICLR)},
  2017.

\bibitem[Malinowski and Fritz(2014)]{malinowski2014multi}
Mateusz Malinowski and Mario Fritz.
\newblock A multi-world approach to question answering about real-world scenes
  based on uncertain input.
\newblock In \emph{Advances in neural information processing systems}, pages
  1682--1690, 2014.

\bibitem[Antol et~al.(2015)Antol, Agrawal, Lu, Mitchell, Batra,
  Lawrence~Zitnick, and Parikh]{antol2015vqa}
Stanislaw Antol, Aishwarya Agrawal, Jiasen Lu, Margaret Mitchell, Dhruv Batra,
  C~Lawrence~Zitnick, and Devi Parikh.
\newblock Vqa: Visual question answering.
\newblock In \emph{Proceedings of the IEEE International Conference on Computer
  Vision (ICCV)}, pages 2425--2433, 2015.

\bibitem[Ravi and Larochelle(2016)]{ravi2016optimization}
Sachin Ravi and Hugo Larochelle.
\newblock Optimization as a model for few-shot learning.
\newblock 2016.

\bibitem[Oreshkin et~al.(2018)Oreshkin, L{\'o}pez, and
  Lacoste]{oreshkin2018tadam}
Boris Oreshkin, Pau~Rodr{\'\i}guez L{\'o}pez, and Alexandre Lacoste.
\newblock Tadam: Task dependent adaptive metric for improved few-shot learning.
\newblock In \emph{Advances in Neural Information Processing Systems}, pages
  721--731, 2018.

\bibitem[Goodfellow et~al.(2014)Goodfellow, Pouget-Abadie, Mirza, Xu,
  Warde-Farley, Ozair, Courville, and Bengio]{Goodfellow2014GenerativeAN}
Ian~J. Goodfellow, Jean Pouget-Abadie, Mehdi Mirza, Bing Xu, David
  Warde-Farley, Sherjil Ozair, Aaron~C. Courville, and Yoshua Bengio.
\newblock Generative adversarial nets.
\newblock In \emph{Advances in Neural Information Processing Systems}, 2014.

\bibitem[Mirza and Osindero(2014)]{Mirza2014ConditionalGA}
Mehdi Mirza and Simon Osindero.
\newblock Conditional generative adversarial nets.
\newblock \emph{CoRR}, abs/1411.1784, 2014.

\bibitem[Odena et~al.(2017)Odena, Olah, and Shlens]{Odena2017ConditionalIS}
Augustus Odena, Christopher Olah, and Jonathon Shlens.
\newblock Conditional image synthesis with auxiliary classifier gans.
\newblock In \emph{International Conference on Machine Learning (ICML)}, 2017.

\bibitem[He et~al.(2016)He, Zhang, Ren, and Sun]{he2016deep}
Kaiming He, Xiangyu Zhang, Shaoqing Ren, and Jian Sun.
\newblock Deep residual learning for image recognition.
\newblock In \emph{Proceedings of the IEEE conference on computer vision and
  pattern recognition (CVPR)}, pages 770--778, 2016.

\bibitem[Gulrajani et~al.(2017)Gulrajani, Ahmed, Arjovsky, Dumoulin, and
  Courville]{Gulrajani2017ImprovedTO}
Ishaan Gulrajani, Faruk Ahmed, Mart{\'i}n Arjovsky, Vincent Dumoulin, and
  Aaron~C. Courville.
\newblock Improved training of wasserstein gans.
\newblock In \emph{Advances in Neural Information Processing Systems}, page
  5769–5779, 2017.

\bibitem[Zhang et~al.(2018{\natexlab{a}})Zhang, Goodfellow, Metaxas, and
  Odena]{Han18sagan}
Han Zhang, Ian~J. Goodfellow, Dimitris~N. Metaxas, and Augustus Odena.
\newblock Self-attention generative adversarial networks.
\newblock \emph{International Conference on Learning Representations (ICLR)},
  2018{\natexlab{a}}.

\bibitem[Brock et~al.(2019)Brock, Donahue, and Simonyan]{Brock2019LargeSG}
Andrew Brock, Jeff Donahue, and Karen Simonyan.
\newblock Large scale gan training for high fidelity natural image synthesis.
\newblock \emph{International Conference on Learning Representations}, 2019.

\bibitem[Parikh et~al.(2016)Parikh, T{\"a}ckstr{\"o}m, Das, and
  Uszkoreit]{parikh2016decomposable}
Ankur~P Parikh, Oscar T{\"a}ckstr{\"o}m, Dipanjan Das, and Jakob Uszkoreit.
\newblock A decomposable attention model for natural language inference.
\newblock \emph{arXiv preprint arXiv:1606.01933}, 2016.

\bibitem[Vaswani et~al.(2017)Vaswani, Shazeer, Parmar, Uszkoreit, Jones, Gomez,
  Kaiser, and Polosukhin]{vaswani2017attention}
Ashish Vaswani, Noam Shazeer, Niki Parmar, Jakob Uszkoreit, Llion Jones,
  Aidan~N Gomez, {\L}ukasz Kaiser, and Illia Polosukhin.
\newblock Attention is all you need.
\newblock In \emph{Advances in neural information processing systems}, pages
  5998--6008, 2017.

\bibitem[Cheng et~al.(2016)Cheng, Dong, and Lapata]{cheng2016long}
Jianpeng Cheng, Li~Dong, and Mirella Lapata.
\newblock Long short-term memory-networks for machine reading.
\newblock \emph{arXiv preprint arXiv:1601.06733}, 2016.

\bibitem[Miyato et~al.(2018)Miyato, Kataoka, Koyama, and
  Yoshida]{miyato2018spectral}
Takeru Miyato, Toshiki Kataoka, Masanori Koyama, and Yuichi Yoshida.
\newblock Spectral normalization for generative adversarial networks.
\newblock \emph{International Conference on Learning Representations (ICLR)},
  2018.

\bibitem[Cho et~al.(2014)Cho, Van~Merri{\"e}nboer, Bahdanau, and
  Bengio]{cho2014properties}
Kyunghyun Cho, Bart Van~Merri{\"e}nboer, Dzmitry Bahdanau, and Yoshua Bengio.
\newblock On the properties of neural machine translation: Encoder-decoder
  approaches.
\newblock \emph{arXiv preprint arXiv:1409.1259}, 2014.

\bibitem[Nair and Hinton(2010)]{nair2010rectified}
Vinod Nair and Geoffrey~E Hinton.
\newblock Rectified linear units improve restricted boltzmann machines.
\newblock In \emph{Proceedings of the 27th international conference on machine
  learning (ICML-10)}, pages 807--814, 2010.

\bibitem[Kingma and Ba(2014)]{kingma2014adam}
Diederik~P Kingma and Jimmy Ba.
\newblock Adam: A method for stochastic optimization.
\newblock \emph{International Conference on Learning Representations (ICLR)},
  2014.

\bibitem[Russakovsky et~al.(2015)Russakovsky, Deng, Su, Krause, Satheesh, Ma,
  Huang, Karpathy, Khosla, Bernstein, et~al.]{russakovsky2015imagenet}
Olga Russakovsky, Jia Deng, Hao Su, Jonathan Krause, Sanjeev Satheesh, Sean Ma,
  Zhiheng Huang, Andrej Karpathy, Aditya Khosla, Michael Bernstein, et~al.
\newblock Imagenet large scale visual recognition challenge.
\newblock \emph{International journal of computer vision}, 115\penalty0
  (3):\penalty0 211--252, 2015.

\bibitem[Jiang et~al.(2018)Jiang, Havaei, Varno, Chartrand, Chapados, and
  Matwin]{jiang2018learning}
Xiang Jiang, Mohammad Havaei, Farshid Varno, Gabriel Chartrand, Nicolas
  Chapados, and Stan Matwin.
\newblock Learning to learn with conditional class dependencies.
\newblock 2018.

\bibitem[Vinyals et~al.(2016)Vinyals, Blundell, Lillicrap, Wierstra,
  et~al.]{vinyals2016matching}
Oriol Vinyals, Charles Blundell, Timothy Lillicrap, Daan Wierstra, et~al.
\newblock Matching networks for one shot learning.
\newblock In \emph{Advances in neural information processing systems}, pages
  3630--3638, 2016.

\bibitem[Krizhevsky(2009)]{cifar10}
Alex Krizhevsky.
\newblock Learning multiple layers of features from tiny images.
\newblock 2009.

\bibitem[Salimans et~al.(2016)Salimans, Goodfellow, Zaremba, Cheung, Radford,
  Chen, and Chen]{salimans2016improved}
Tim Salimans, Ian Goodfellow, Wojciech Zaremba, Vicki Cheung, Alec Radford,
  Xi~Chen, and Xi~Chen.
\newblock Improved techniques for training gans.
\newblock In \emph{Advances in Neural Information Processing Systems}, 2016.

\bibitem[Heusel et~al.(2017)Heusel, Ramsauer, Unterthiner, Nessler, and
  Hochreiter]{Heusel2017GANsTB}
Martin Heusel, Hubert Ramsauer, Thomas Unterthiner, Bernhard Nessler, and Sepp
  Hochreiter.
\newblock Gans trained by a two time-scale update rule converge to a local nash
  equilibrium.
\newblock In \emph{Advances in Neural Information Processing Systems}, 2017.

\bibitem[Paszke et~al.(2017)Paszke, Gross, Chintala, Chanan, Yang, DeVito, Lin,
  Desmaison, Antiga, and Lerer]{paszke2017automatic}
Adam Paszke, Sam Gross, Soumith Chintala, Gregory Chanan, Edward Yang, Zachary
  DeVito, Zeming Lin, Alban Desmaison, Luca Antiga, and Adam Lerer.
\newblock Automatic differentiation in {PyTorch}.
\newblock In \emph{NeurIPS Autodiff Workshop}, 2017.

\bibitem[Abadi et~al.(2015)Abadi, Agarwal, Barham, Brevdo, Chen, Citro,
  Corrado, Davis, Dean, Devin, Ghemawat, Goodfellow, Harp, Irving, Isard, Jia,
  Jozefowicz, Kaiser, Kudlur, Levenberg, Man\'{e}, Monga, Moore, Murray, Olah,
  Schuster, Shlens, Steiner, Sutskever, Talwar, Tucker, Vanhoucke, Vasudevan,
  Vi\'{e}gas, Vinyals, Warden, Wattenberg, Wicke, Yu, and
  Zheng]{tensorflow2015-whitepaper}
Mart\'{\i}n Abadi, Ashish Agarwal, Paul Barham, Eugene Brevdo, Zhifeng Chen,
  Craig Citro, Greg~S. Corrado, Andy Davis, Jeffrey Dean, Matthieu Devin,
  Sanjay Ghemawat, Ian Goodfellow, Andrew Harp, Geoffrey Irving, Michael Isard,
  Yangqing Jia, Rafal Jozefowicz, Lukasz Kaiser, Manjunath Kudlur, Josh
  Levenberg, Dan Man\'{e}, Rajat Monga, Sherry Moore, Derek Murray, Chris Olah,
  Mike Schuster, Jonathon Shlens, Benoit Steiner, Ilya Sutskever, Kunal Talwar,
  Paul Tucker, Vincent Vanhoucke, Vijay Vasudevan, Fernanda Vi\'{e}gas, Oriol
  Vinyals, Pete Warden, Martin Wattenberg, Martin Wicke, Yuan Yu, and Xiaoqiang
  Zheng.
\newblock {TensorFlow}: Large-scale machine learning on heterogeneous systems,
  2015.
\newblock URL \url{http://tensorflow.org/}.
\newblock Software available from tensorflow.org.

\bibitem[Szegedy et~al.(2016)Szegedy, Vanhoucke, Ioffe, Shlens, and
  Wojna]{szegedy2016rethinking}
Christian Szegedy, Vincent Vanhoucke, Sergey Ioffe, Jon Shlens, and Zbigniew
  Wojna.
\newblock Rethinking the inception architecture for computer vision.
\newblock In \emph{Proceedings of the IEEE conference on computer vision and
  pattern recognition}, pages 2818--2826, 2016.

\bibitem[Barratt and Sharma(2018)]{Barratt2018ANO}
Shane Barratt and Rishi~Kant Sharma.
\newblock A note on the inception score.
\newblock \emph{CoRR}, abs/1801.01973, 2018.

\bibitem[Ravuri and Vinyals(2019)]{ravuri2019cas}
Suman Ravuri and Oriol Vinyals.
\newblock Classification accuracy score for conditional generative models.
\newblock \emph{arXiv preprint arXiv:1905.10887}, 2019.

\bibitem[Zhang et~al.(2019)Zhang, Dauphin, and Ma]{zhang2018fixup}
Hongyi Zhang, Yann~N. Dauphin, and Tengyu Ma.
\newblock Fixup initialization: Residual learning without normalization.
\newblock In \emph{International Conference on Learning Representations}, 2019.

\bibitem[Zhang et~al.(2018{\natexlab{b}})Zhang, Cisse, Dauphin, and
  Lopez-Paz]{zhang2018mixup}
Hongyi Zhang, Moustapha Cisse, Yann~N. Dauphin, and David Lopez-Paz.
\newblock mixup: Beyond empirical risk minimization.
\newblock In \emph{International Conference on Learning Representations},
  2018{\natexlab{b}}.
\newblock URL \url{https://openreview.net/forum?id=r1Ddp1-Rb}.

\end{thebibliography}
\end{document}